# Efficient Design of Neural Networks with Random Weights

Ajay M. Patrikar
ajay.m.patrikar@gmail.com

*Abstract*—Single layer feedforward networks with random weights are known for their non-iterative and fast training algorithms and are successful in a variety of classification and regression problems. A major drawback of these networks is that they require a large number of hidden units. In this paper, we propose a technique to reduce the number of hidden units substantially without affecting the accuracy of the networks significantly. We introduce the concept of primary and secondary hidden units. The weights for the primary hidden units are chosen randomly while the secondary hidden units are derived using pairwise combinations of the primary hidden units. Using this technique, we show that the number of hidden units can be reduced by at least one order of magnitude. We experimentally show that this technique leads to significant drop in computations at inference time and has only a minor impact on network accuracy. A huge reduction in computations is possible if slightly lower accuracy is acceptable.

*Keywords— Machine learning, feedforward neural networks, neural networks with random weights, random vector functional link networks, extreme learning machines*

## I. Introduction

Single layer feedforward networks with random weights were first proposed in the early nineties [1-4]. In the last two decades they have been successfully applied to a large number of pattern classification and regression problems [5]. These networks have often been referred to as random vector functional link (RVFL) networks [6-10] or extreme learning machines (ELM) [11-12] in the literature. In this paper, we will refer to them as neural networks with random weights (NNRW). For these networks, the weights between the input and hidden layer are assigned randomly and are not trained. Weights between the hidden and output layer are obtained analytically using non-iterative training algorithms. These algorithms [5] are known to be much faster than the conventional neural networks, which depend on iterative training based on error backpropagation.

A major drawback of NNRW is that a large number of hidden units are required by these networks to achieve good accuracy. This can result in a longer running time during inference, which can limit their use on platforms with limited computational power such as embedded systems, Internet of Things, smartphones, drones, etc. Presently, machine learning algorithms are increasingly being adopted on such platforms, emphasizing the need for efficient machine learning models.

There have been several attempts to reduce the number of hidden units reported in the literature [11-19]. These methods often depend on incrementally adding or pruning hidden units in the network. In this paper, we take a very different approach by altering the basic design of the network. We introduce the concept of primary and secondary hidden units.

The purpose of the primary hidden units is to take a weighed sum of their inputs. The purpose of the secondary hidden units is to take a pairwise sum of the outputs the primary hidden units and apply a nonlinear activation function. We show that secondary hidden units are equivalent to additional hidden units without the burden of extra computations. The proposed design leads to increase in the number of tunable parameters without increasing the number of random weights. This results in at least one order of magnitude drop in the number of primary hidden units and corresponding drop in total computations. We experimentally show that the proposed design has very small impact on network accuracy. On the other hand, a huge reduction in the number of hidden units is possible if a slight drop in accuracy is tolerable.

The paper is organized as follows. Section II introduces the efficient design of NNRW. The design is analyzed to show that it is equivalent to addition of extra hidden units without the extra computations. In Section III, experimental results are presented on a number of benchmark machine learning problems. Section IV gives a summary and conclusions.

## II. Efficient Design of NNRW

The basic design of a single layer NNRW [5] is shown in Fig. 1. Each hidden unit takes a weighted sum of its inputs and applies a nonlinear activation function. The weights between the input layer and hidden layer are assigned randomly and not trained. The weights between the hidden layer and the output layer are tunable and are learned from the training data. In some implementations of NNRW, there are direct connections between the input and output layer (e.g. RVFL networks). While the efficient design proposed in this paper can accommodate such variations, we focus our analysis only on the architecture shown in Fig. 1.

To obtain efficient design of NNRW we modify the hidden layer to include primary and secondary hidden units as shown in Fig. 2. The primary hidden unit takes a weighted sum of its inputs. There are no weights associated with the links between the primary and secondary hidden units. The primary units and secondary units are sparsely connected, i.e., each secondary unit connects to only two primary units. The function of the secondary unit is to take a sum of outputs of the attached pair of primary units and apply an activation function.

Let $\vec{x}$ be the input feature vector, $\vec{a}_i$ be the random weight vector, and $b_i$ be the random bias term associated with the $i$th primary hidden unit. The output of the primary hidden unit is

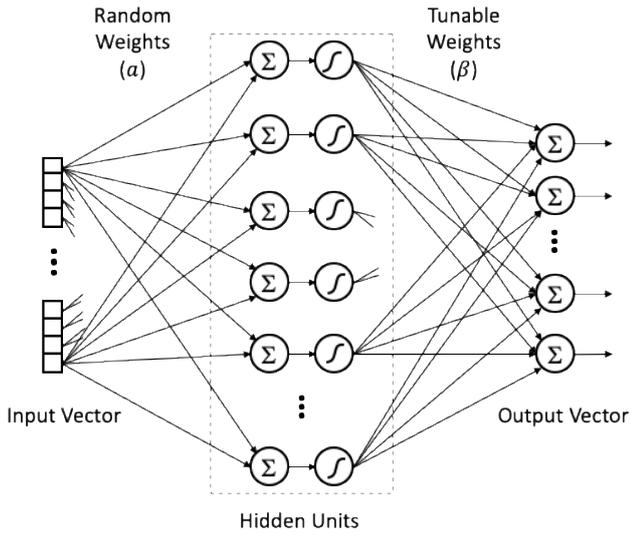

Fig. 1. Basic design of NNRW

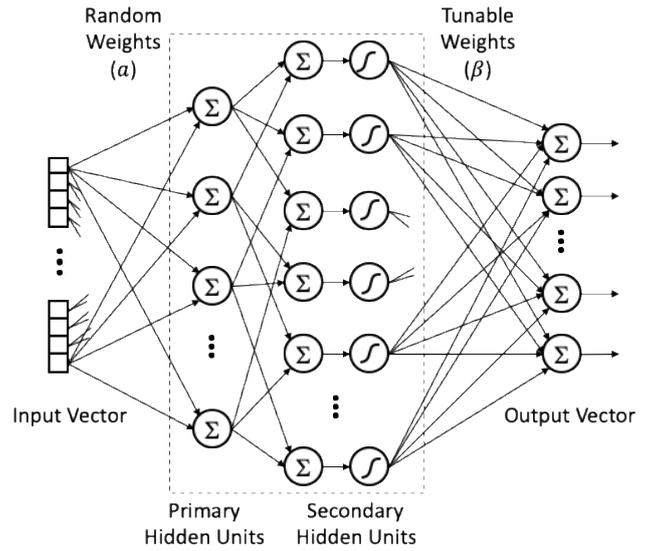

Fig. 2. Efficient design of NNRW consisting of primary and secondary hidden units

given by

$$p_i = \vec{a}_i \cdot \vec{x} + b_i \quad (1)$$

The secondary hidden unit takes a pairwise sum of the primary hidden units and applies an activation function $g(\ )$. Assuming primary hidden units $i$ and $j$ have connections to the secondary hidden unit $k$, the output of the secondary unit $s_k$ is expressed as

$$s_k = g(p_i + p_j) = g((\vec{a}_i + \vec{a}_j) \cdot \vec{x} + (b_i + b_j)) \quad (2)$$

This is equivalent to introducing a hidden unit in which the random weights are obtained by adding two sets of random weights associated with the primary hidden units. In NNRW, the random weights are drawn from certain probability distributions, e.g. uniform distribution, normal distribution, etc. The probability distribution of a sum of independent random variables is the convolution of the probability distributions of those random variables [28]. The convolution of two normal distributions is also normal except that the mean and variance are different. For example, if the random weights of the primary hidden units are drawn from normal distribution $N(\mu, \sigma^2)$, where $\mu$ is the mean and $\sigma^2$ is the variance, it is equivalent to having a secondary hidden unit whose random weights are drawn from distribution $N(2\mu, 2\sigma^2)$. In case of random weights drawn from uniform distributions, their sums result in random weights drawn from a triangular distribution. If there are $P$ primary units, the maximum number of secondary units, $M$, derived from them are:

$$M = \binom{P}{2} = P(P-1)/2. \quad (3)$$

However, it is not necessary to use all the combinations of primary hidden units, which then leads to a smaller number of secondary units. In our experiments, we connect a primary unit $i$ to another primary unit $(i + \tau)$, where $\tau$ is a positive integer (i.e. $\tau \geq 1$) until all the combinations are exhausted. The relationships between the number of primary and secondary units is approximately given by:

$$M \approx \frac{P(P-1)}{2\tau} \approx \frac{P^2}{2\tau} \quad (4)$$

In some of our experiments we set the number of secondary units approximately about ten times the number of primary units (i.e. $P = M/10$). This leads to the following equation:

$$\tau = M/200. \quad (5)$$

To train the tunable weights we extend the training algorithm for the basic NNRW. We construct a vector $\vec{h}(\vec{x})$ consisting of output of secondary units:

$$\vec{h}(\vec{x}) = [s_1(\vec{x}), s_2(\vec{x}), \ldots, s_M(\vec{x})]. \quad (6)$$

The dimension of this vector is $M$, where $M$ is the number of secondary units. We define the output function for each class to be

$$f^n(\vec{x}) = \vec{h}(\vec{x}) \cdot \vec{\beta}^n \quad (7)$$

where $\vec{\beta}^n = [w_1^n, w_2^n, \ldots, w_M^n]^T$ is a vector of the output weights for the $n$th class. Our goal is to determine the output weights $\vec{\beta}^n$ for each class.

Given $L$ training samples $\{(\vec{x}_i, \vec{t}_i)\}_{i=1}^{L}$, we seek a solution to the following learning problem:

$$H\,\beta = T \quad (8)$$

where $T = [\vec{t}_1, \ldots, \vec{t}_L]^T$ are target labels, $H = [\vec{h}(\vec{x}_1), \ldots, \vec{h}(\vec{x}_L)]^T$ is a matrix consisting of secondary hidden unit output vectors, and $\beta = [\vec{\beta}^1, \ldots, \vec{\beta}^Q]^T$ is the output weight matrix. There are $Q$ classes. The output weights $\beta$ can be calculated as follows:

$$\beta = H^\dagger T \quad (9)$$

where $H^\dagger$ is the Moore-Penrose generalized inverse of matrix $H$. There are several methods for calculation of $H^\dagger$. These include the orthogonal projection method, orthogonalization method, iterative method, and singular value decomposition

[20-21]. Another alternative is to use ridge regression [22-23] for which a solution is given by

$$\boldsymbol{\beta} = \boldsymbol{H}(\boldsymbol{H}^T\boldsymbol{H} + \lambda \boldsymbol{I})^{-1}\boldsymbol{T} \qquad (10)$$

where $\boldsymbol{I}$ is the $M \times M$ identity matrix and $\lambda$ is a tunable parameter. We use this method in our experiments with $\boldsymbol{\lambda}$ set to 0.01. For a network of $N$ inputs, $P$ primary hidden units, $M$ secondary hidden units, and $Q$ outputs, the number of multiply-and-accumulate arithmetic operations during inference is approximately

$$NP + MQ \qquad (11)$$

if the bias terms are ignored. We use this formula to compare network computations. For the basic design of NNRW, (11) still applies, except that $P$ equals $M$. Hence the number of arithmetic operations for the basic NNRW is approximately

$$(N + Q)M \qquad (12)$$

In NNRW, the random projection performed by the hidden layer usually does not contain any information specific to the classification or regression problem that the network is trying to solve. However, there is significant amount of computation associated with the random projection step. With the creation of primary and secondary units we limit the number of random parameters and associated computations. At the same time, we increase the number of tunable weights $\boldsymbol{\beta}$ by introducing secondary hidden units. Increase in tunable weights often leads to better performance, until overfitting causes the performance to deteriorate.

### III. EXPERIMENTAL RESULTS

In this section, we describe experiments on three benchmark machine learning problems. In each case, the proposed efficient version of NNRW (Fig. 2) is compared with the baseline model (Fig. 1), which uses the basic design. In all experiments we use the sigmoid activation function.

When $\tau$ is set to 1, it leads to a model in which the number of primary units $P \approx \sqrt{2M}$, which is a very small number compared to $M$. In our simulations, we observed that $\tau = 1$ can lead to a reduced accuracy, most likely because of repeated use of primary unit random vectors in calculation of secondary hidden unit outputs. To determine $\tau$ we use the following procedure. We first train the baseline classifier and determine the number of hidden units required to obtain the highest accuracy. We then set the number of secondary units $M$ to that number. The $\tau$ parameter is then calculated using (5).

#### A. Results of the SatImage Problem

NNRW classifiers were trained for the Landsat satellite image (SatImage) problem from the Statlog [25] collection. This problem contains 36 attributes, six classes, 4,435 training samples, and 2,000 test samples. Twenty-five trials were conducted with different random initializations and the average classification accuracy was calculated. Fig. 3 shows the average classification accuracy as a function of the number of hidden units when the $\tau$ parameter was set to 4. There are two sets of numbers along the x-axis. The bottom row contains the number of primary units for the efficient NNRW. The top row contains the number of secondary units for the efficient NNRW as well as the number of hidden units for the baseline NNRW.

The best accuracy of 90.40% was obtained by the efficient NNRW with 90 primary hidden units which corresponds to 968 secondary hidden units for $\tau$ parameter set at 4. The best accuracy obtained by the baseline NNRW was 90.30% with 968 hidden units. Thus, the efficient NNRW requires less than one tenth of the primary hidden units as compared to the baseline NNRW. The top performing efficient NNRW requires 77% fewer computations compared to the top performing baseline NNRW as calculated from (11) and (12). In this case, the accuracy of the top performing efficient NNRW is slightly higher than the top performing baseline version.

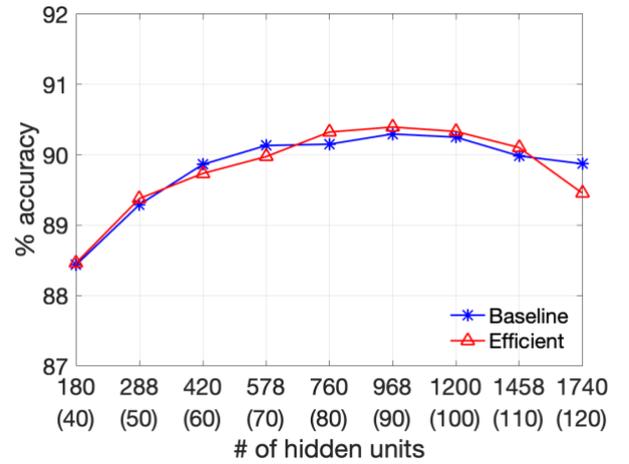

Fig. 3. Results of SatImage classification problem for $\tau = 4$

The results of the experiment when $\tau$ is set to 1 are shown in Fig. 4. In this case there is a slight drop in accuracy. The best accuracy obtained by efficient NNRW was 90.17% with 40 hidden units. This is lower than 90.39% accuracy obtained by the baseline version with 780 hidden units. The efficient model requires 81% fewer computations.

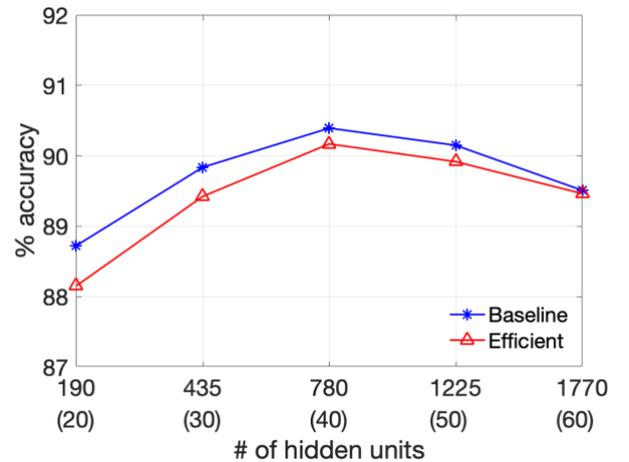

Fig. 4. Results of SatImage classification problem for $\tau = 1$

#### B. Results of the UCI Letter Recognition Problem

The UCI letter recognition problem [25] contains 16 attributes and 26 classes. The data consist of 20,000 samples. For each trial, the training data set and test data set are randomly generated from the overall database. 13,333 samples were used for training and 6,667 samples were used for testing. Twenty-five trials were conducted with different random initializations as well as data partitions, and the average classification accuracy was calculated. Fig. 5 shows the average classification accuracy as a function of the number of hidden units when the $\tau$ parameter was set to 28.

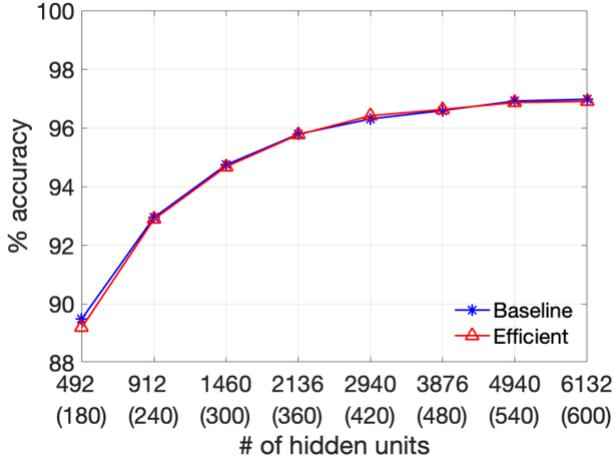

Fig. 5. Results of the UCI letter recognition problem for $\tau = 28$

The best accuracy of 96.90% was obtained by efficient NNRW with 600 primary hidden units which corresponds to 6,132 secondary hidden units for $\tau$ parameter set at 28. The best accuracy obtained by baseline NNRW was 96.98% with 6,132 hidden units. The efficient NNRW requires less than one-tenth of the primary hidden units as compared to the baseline NNRW. The top performing efficient NNRW requires 34% fewer computations compared to the top performing baseline NNRW as calculated from (11) and (12). There is a slight reduction in accuracy associated with the use of efficient NNRW.

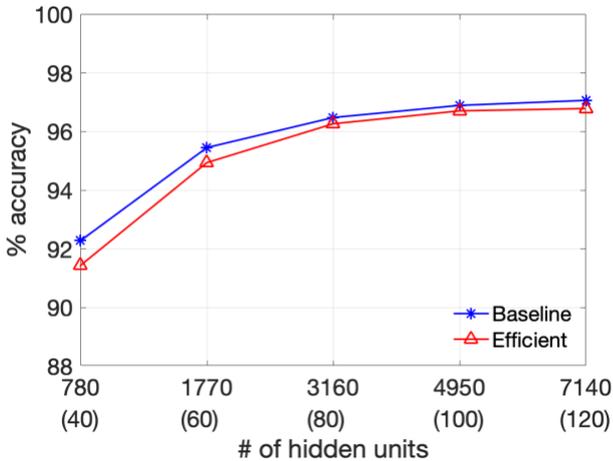

Fig. 6. Results of the UCI letter recognition problem for $\tau = 1$

The results of the experiment when $\tau$ is set to 1 are shown in Fig. 6. The best accuracy obtained by the efficient NNRW was 96.78% with 120 primary hidden units. This is lower than 97.06% accuracy obtained by the baseline version with 7140 hidden units. The efficient model requires 37% fewer computations.

## C. Results of the MNIST Classification Problem

MNIST is a benchmark problem for handwritten digit recognition [26]. The problem consists of 10 classes, 60,000 training images, and 10,000 test images. The dimensionality of images is 28x28 pixels. We used the original MNIST dataset without any distortions. Thus, the dimensionality of the feature vector was 784. For this problem, we made use of shaped input weights to initialize hidden layer weights as described in [22, 27] which are known to provide better accuracy. Fig. 7 shows the average classification accuracy over twenty-five trials as a function of the number of hidden units when the $\tau$ parameter was set to 90.

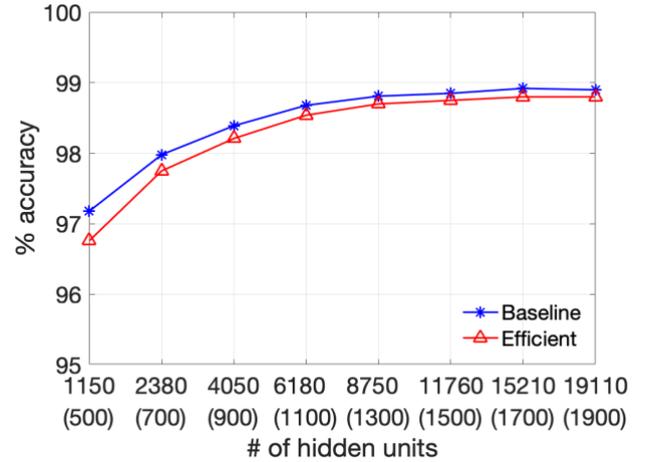

Fig. 7. Results of the MNIST classification problem for $\tau = 90$

The best accuracy of 98.80% was obtained by the efficient NNRW with 1,900 primary hidden units which corresponds to 19,110 secondary hidden units for $\tau$ parameter set at 90. The best accuracy obtained by the baseline NNRW was 98.90% with 19,110 hidden units. The efficient NNRW requires about one-tenth of the primary hidden units as compared to the baseline NNRW. The top performing efficient NNRW requires 88% fewer computations compared to the top performing baseline NNRW as calculated from (11) and (12). There is a minor reduction in accuracy associated with the use of efficient NNRW. This experiment shows that the proposed method also works for shaped input weights which are not random.

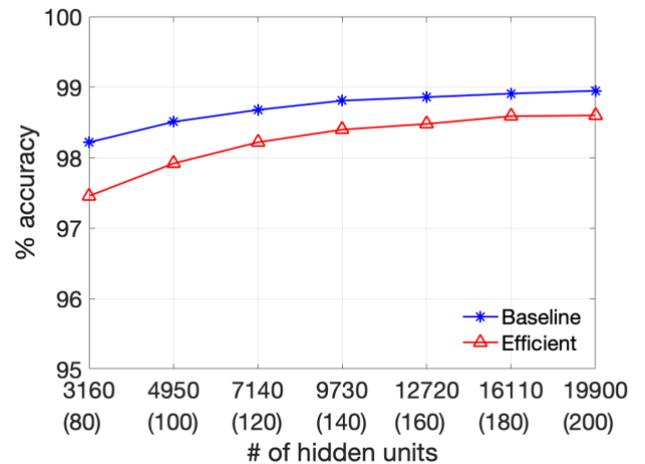

Fig. 8. Results of the MNIST classification problem for $\tau = 1$

The results of the experiment when τ is set to 1 are shown in Fig. 8. The best accuracy obtained by the efficient NNRW was 98.6% with 200 primary hidden units. This is lower than 98.94% accuracy obtained by the baseline version with 19,900 hidden units. The efficient model requires 97% fewer computations.

## IV. CONCLUSION

We have shown that the proposed efficient design of NNRW results in significant reduction in the number of hidden units and associated computations. Our experiments show that the classification accuracy of the efficient NNRW is very close to the basic design for settings in which the number of primary hidden units are one-tenth of the baseline version. It is certainly possible to reduce the computations greatly if a small reduction in accuracy is tolerable. The proposed design results in smaller models in terms of both memory and computations making them suitable for computational platforms with limited resources.